\documentclass[a4paper, 10pt, conference]{ieeeconf}

\IEEEoverridecommandlockouts   
\usepackage{graphicx}
\usepackage{float}
\usepackage{mathptmx}
\usepackage{times}
\usepackage{amsmath}
\usepackage{amssymb}
\usepackage{url}
\usepackage[colorlinks=true, allcolors=blue]{hyperref}
\usepackage{paralist} 
\usepackage{subcaption}
\usepackage{tikz}
\usetikzlibrary{arrows}

\title{\LARGE \bf
Understanding aesthetics in photography using\\deep convolutional neural networks
}

\author{ Maciej Suchecki, Tomasz Trzcinski\\
  Faculty of Electronics and Information Technology\\
  Warsaw University of Technology\\
  Nowowiejska 15/19, 00-665 Warsaw, Poland\\
  {\tt\small m.suchecki@stud.elka.pw.edu.pl}, {\tt\small t.trzcinski@ii.pw.edu.pl}
}

\begin{document}

\maketitle
\thispagestyle{empty}
\pagestyle{empty}

\begin{abstract}
  Evaluating aesthetic value of digital photographs is a challenging task, mainly due to numerous factors that need to be taken into account and subjective manner of this process. In this paper, we propose to approach this problem using deep convolutional neural networks. Using a dataset of over 1.7 million photos collected from Flickr, we train and evaluate a deep learning model whose goal is to classify input images by analysing their aesthetic value. The result of this work is a publicly available Web-based application that can be used in several real-life applications, e.g. to improve the workflow of professional photographers by pre-selecting the best photos.
\end{abstract}

\section{Introduction}
\label{sec:introduction}
Predicting aesthetic value of photographs is a~challenging task for a~computer-based system. Humans experience a~lot of difficulties when explaining why a~given picture is perceived as aesthetically pleasing. This is why it does not seem to be possible to solve this challenge by defining a set of rules (such as \textit{if a photo has a lot of blue color it is considered beautiful}). Instead, we believe that this problem can be addressed by referring to a crowd-sourced dataset of photographs with corresponding popularity score which we treat as aesthetic metric proxy. Training a machine learning algorithm using this dataset appears to be a more practical approach to estimating the aesthetic value of a photograph and we follow this methodology here.

More precisely, in this paper, we assess the aesthetic value of a~photograph using only the values of its pixels. Building up on the successful applications of deep convolutional neural networks in other related domains, such as image recognition~\cite{lecun1998,ilsvrc15,imagenet2012}, we propose to use this approach to address this problem. Our method is strongly inspired by previous research on similar problems, which concluded that deep convolutional neural networks perform well in such cases~\cite{flickrpopularity2014,gelli2015,flickrstyle2013,deng2016}.

One could imagine a~wide variety of applications for a~system solving the problem stated in the paper. First of all, such a~system can significantly improve the workflow of every photographer. By preselecting or suggesting the best photos from a~defined set we save a~lot of time, storage space and network traffic. Usefulness of the system could be furthermore improved by combining it with some means of detecting similar photos. That results in easy removal of duplicates, which saves photographer the time that he would spend on selecting the best frame out of a~set of similar ones. To provide another example, one could even imagine camera or post processing software using such system to suggest and perform automatic image enhancements, like exposure compensation or even cropping and framing a~photograph.

\noindent Our paper provides the following contributions:
\begin{compactitem}
  \item a machine learning system capable of automatic assessment of aesthetic value of digital photographs basing only on the image content,
  \item publicly available web interface that allows anyone to test the system on their own photos,
  \item dataset, published online, consisting of images and labels that we use to train and evaluate our system.
\end{compactitem}

The remainder of this paper is organized in the following way: firstly, we show a brief overview of the related work. Then, after more formal problem definition, we describe our method. The next part contains thorough description of experiments conducted in order to evaluate and improve the performance of the system. Later, we briefly describe the implementation of our web application. Lastly, we conclude the paper, mentioning also future research possibilities.

\section{Related work}
\label{sec:relatedwork}
Training machine learning systems on photos seems to be sensible in the wake of ever-growing popularity of online photo sharing websites, like Flickr or Instagram. The ubiquity of image data available nowadays is not only making machine learning systems easier to train, but also increasing their significance. The growing volume of image data is constantly making it harder to maintain. Machine learning systems doing for example image classification or face recognition are making this data more useful to us.

While so far there is little research on the topic of this paper, numerous papers tackling similar problems are published. One example is provided by an attempt to address the problem of popularity prediction for online images available on Flickr published in 2014 by Khosla \textit{et al.}~\cite{flickrpopularity2014}. The authors use a dataset of over 2 milion images from the mentioned service to extract various visual features from them and train a set of Support Vector Machines on the resulting data. In their research, they prove that pre-trained deep learning convolutional neural networks are the best extractors of data for SVMs from the images. This fact encourages us to evaluate convolutional neural networks in our paper. In the discussed article, the authors use the 'ImageNet network'~\cite{imagenet2012} trained on images from the ImageNet~\cite{imagenet2009} challenge. Concretely, they extract features from the fully connected layer before the classification layer, which outputs a vector of size 4096.

Another paper tackling the problem of popularity prediction for online images was published in 2015 by Gelli \textit{et al.}~\cite{gelli2015}. In their approach, they explore additional cues that could help to predict the popularity of a photo. Similarly to Khosla \textit{et al.}, the proposed model is trained on object, context and user features. Object features are extracted by passing every image through a deep convolutional neural network (with 16 layers), resulting in 4,096-dimensional representation from the 7th fully connected layer. Examples of context features in this case are tags, descriptions and location data of images. User features contain data like the mean views of the images of the photo author. However, compared to Khosla \textit{et al.}, the authors propose to use three new context features and -- most importantly -- visual sentiment features. To discover which visual emotions are associated with a particular image, a visual sentiment concept classification is performed based on the Visual Sentiment Ontology. This additional input allows their model to perform better than that proposed by Khosla \textit{et al.}, especially when classifying images coming from the same user. 

A paper published by Karayev \textit{et al.} in 2013 provides another interesting and related approach~\cite{flickrstyle2013}. The authors try to train a classifier which could recognize a visual style that characterizes a given painting or photograph. They use different datasets, including photos from Flickr, like in the previous example. This problem seems quite related to the task that we tackle in this particular paper. Again, deep learning convolutional neural network used in a similar way like in the paragraph above, proved to be the best method of all of the tested ones. This tells us that deep convolutional architectures are the current state of the art in image-related machine learning problems.

Research which is the greatly related to our problem was published by Deng \textit{et al.} in 2016~\cite{deng2016}. In the paper, the authors summarize different state-of-the-art techniques used today in the assessment of image aesthetic quality. In addition to that, they also publish results of the evaluation of those methods on various datasets. The best results are obtained with deep learning models, which is consistent with research discussed above.. That further justifies the usage of such models in our research.

While the next example is not a published research, it is worth mentioning because of high correlation to the topic of this paper. An article~\cite{eyeem2016}, published on NVIDIA Developer Blog in 2016, is discussing exactly the same topic as in our paper. The article describes that the solution is also based on a convolutional neural network. However, the research is not reproducible because authors do not publish the dataset nor the implementation details. Moreover, they use a private dataset of photographs manually curated by award-winning photographers, which is expected to provide training data of very high quality, hence greatly improving the performance of the trained model.

\section{Method}
\label{sec:method}
In the following sections, we describe our proposal in detail. First section contains formal problem definition, whereas in the next one we describe the dataset.

\subsection{Problem definition}
In order to simplify the problem, we define it as a binary classification task. That means that our dataset contains photos classified either as aesthetically pleasing or not. We define our objective as a task of photo classification as aesthetically pleasing or not, using only visual information.

Formally, we define a set of $N$ samples, $\{x_i, l_i\}$ where $x_i$ is a photo and $l_i \in \{0,1\}$ is a corresponding aesthetic label. The label is equal to 1, if the system perceives the photo as aesthetically pleasing, or equal to 0 in the opposite case. Therefore, we aim to train to classifier $C$ that given a sample (photograph) $x_i \in X$ predicts a label $\hat{l_i} = f_C(x_i, \theta)$, where $\theta$ is a set of parameters of a classifier $C$. Going further, our concrete goal is to minimize the loss function defined below:

\vspace{-2mm}

\begin{equation*}
  \min_{\theta} \, \sum_{i=1}^N \, \left[ l_{i} \, \log f_C(x_i, \theta) + (1-l_{i}) \, \log (1-f_C(x_i, \theta)) \right]
\end{equation*}

\vspace{2mm}

The function defined above is widely known as a multinominal logistic loss function, which is an equivalent to the cross entropy loss function.

\subsection{Dataset}
\label{sec:dataset}
Our dataset consists of 1.7 milion photos downloaded from \textit{Flickr} -- a popular image and video hosting service. In order to classify the photos, we use various metadata associated with them. Basing on work of Khosla \textit{et al.}, we use number of views for a given photo as a main measure of image quality. Furthermore, as was shown~\cite{popularity2010} in 2010, we know that visual media on the Internet tend to receive more views over time. To suppress this effect, we normalize each metric using the upload date, i.e. each value is divided by number of days since the photo was uploaded.

Concretely, our aesthetics score is defined by the following equation:
\begin{equation*}
  score = log_{2}(n_{views} + 1 / n_{days} + 1)
\end{equation*}
where:
\begin{description}
  \setlength{\itemsep}{0pt}
  \setlength{\parskip}{0pt}
  \setlength{\parsep}{0pt} 
  \item[$score$] is photo aesthetics score,
  \item[$n_{views}$] is number of views for a given photo,
  \item[$n_{days}$] is number of days since the given photo was uploaded.
\end{description}

\begin{figure}[H]
  \begin{center}
    \includegraphics[width=0.53\textwidth]{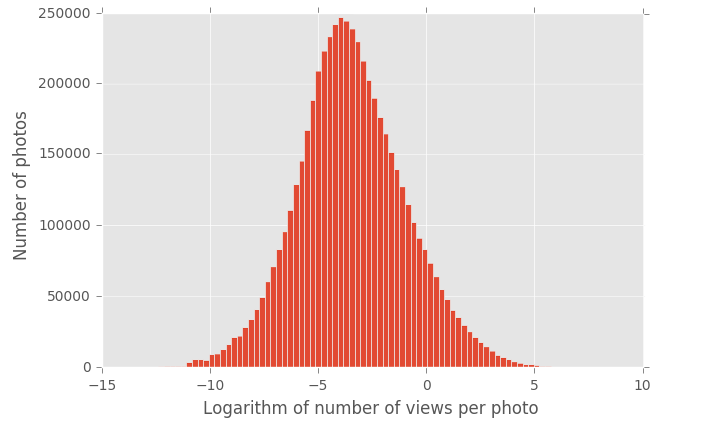}
  \end{center}
  \vspace{-2mm}
  \caption{\label{fig:score_distribution} Distribution of the logarithm of number of views per photo, normalized using upload date for a given photo.}
\end{figure}

This results in distribution plotted on Figure~\ref{fig:score_distribution}. The score defined above is computer for all of the photos in the dataset -- the computed value is used to sort the examples from the dataset. Photos with the biggest value of the score are classified as aesthetically pleasing, whereas photos with the lowest value are classified as not pleasing. Moreover, we focus on top and bottom 20\% of the dataset -- our goal is to distinguish very good photos from those of low quality. That amounts to exactly 513 382 photos in the training dataset and 85 562 photos in the test dataset.

In order to determine if our score is in fact correlated to the quality of the photo, we perform a sanity check. To do so, we sort the photos by their aesthetics score calculated using the formula defined above. Next, we manually assess the aesthetic value of the 100 best and 100 worst photos, i.e. with the highest and lowest score. We do so in order to ensure that our dataset contains valuable examples.

\begin{figure}[H]
  \begin{center}
    \begin{subfigure}[b]{0.48\textwidth}
      \includegraphics[width=\textwidth]{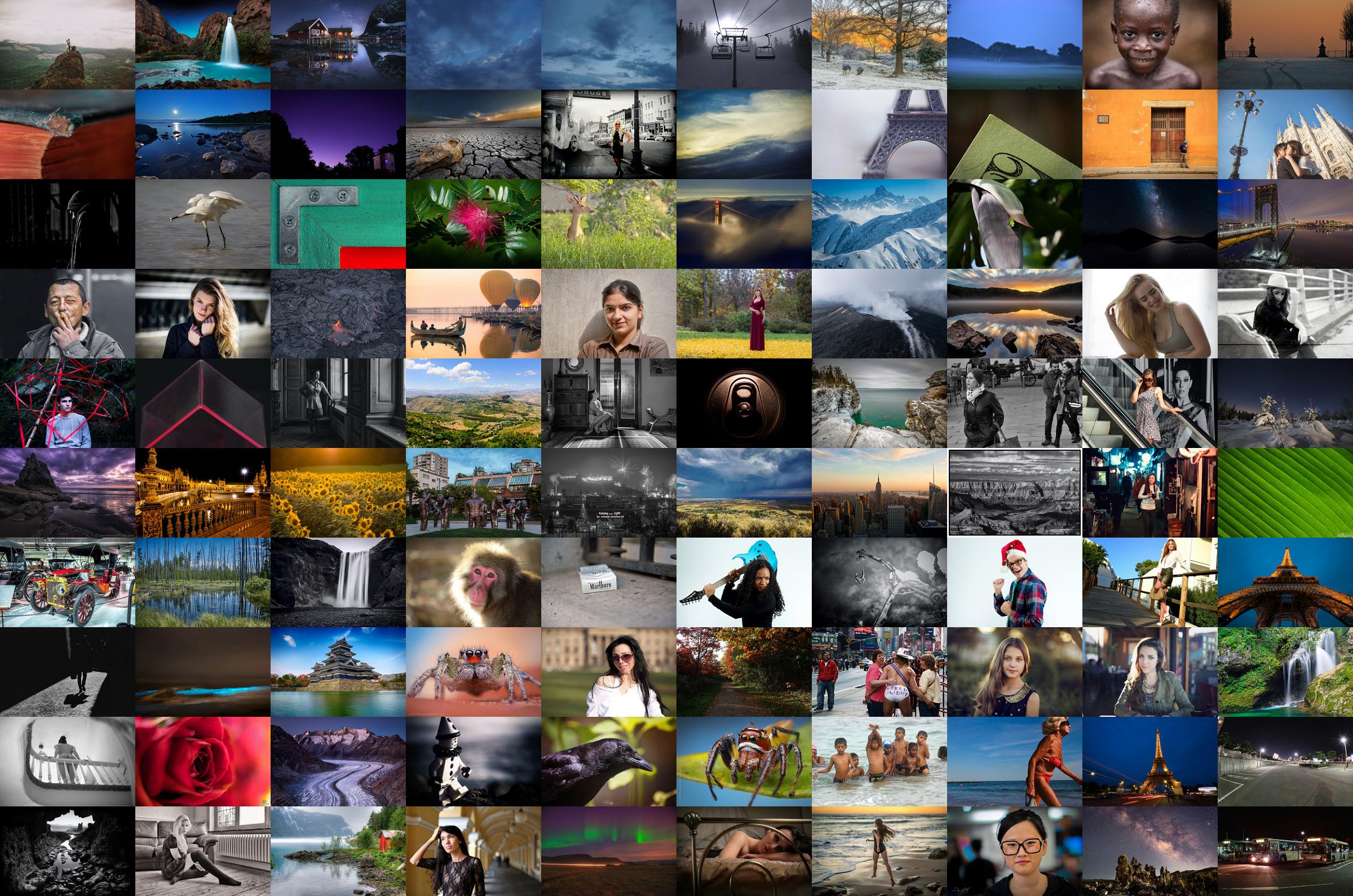}
      \caption{Most viewed photos.}
    \end{subfigure}
  \end{center}
  \begin{center}
    \begin{subfigure}[b]{0.48\textwidth}
      \includegraphics[width=\textwidth]{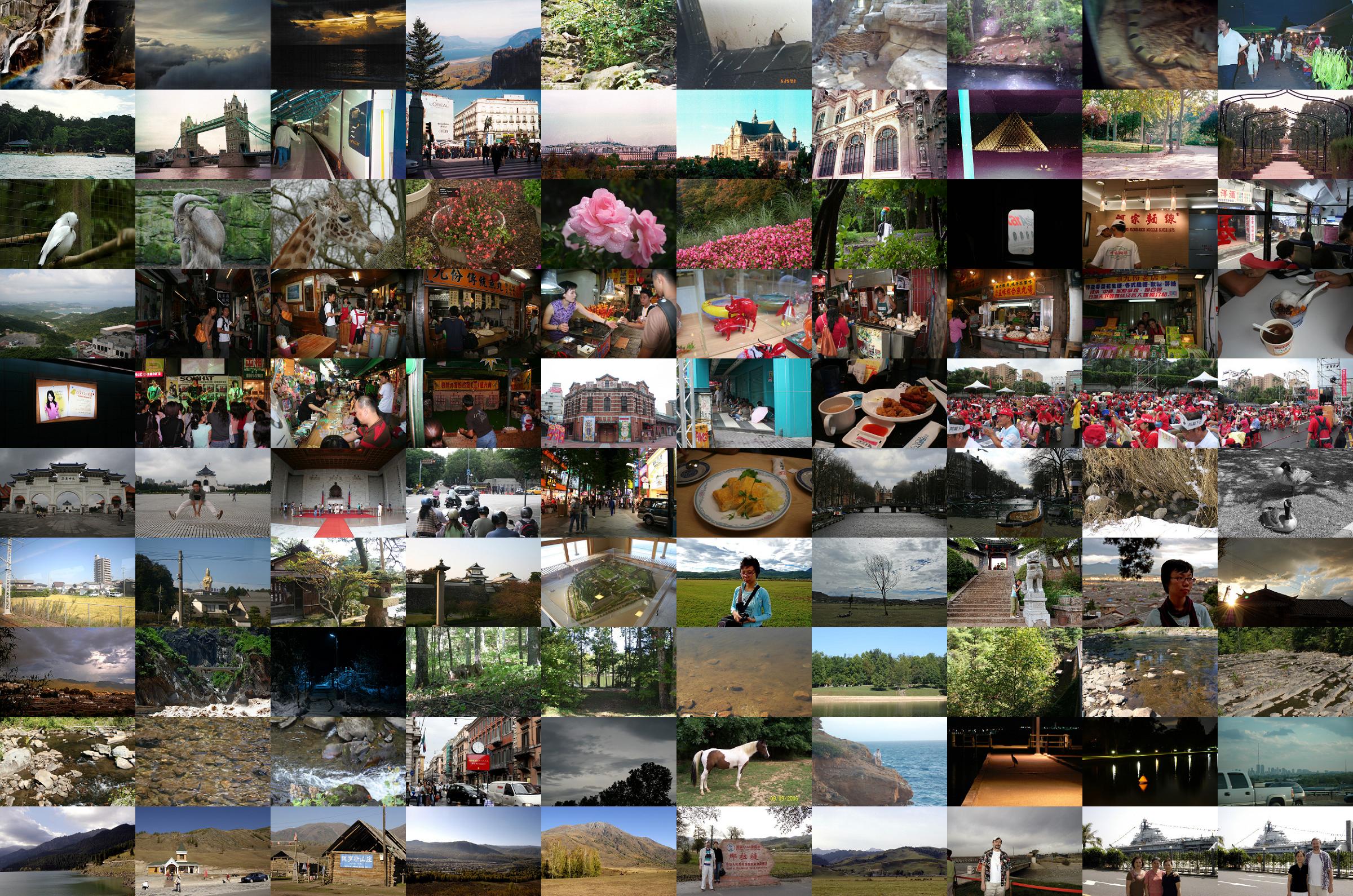}
      \caption{Least viewed photos.}
    \end{subfigure}
  \end{center}
  \vspace{-2mm}
  \caption{ \label{fig:mosaics} 
  Sanity check -- two sets of 100 thumbnails with the (a)~highest and (b)~lowest score from the training set.}
\end{figure} 

Manual examination of the photos on Figure~\ref{fig:mosaics} proves that the photos are classified properly, i.e. the positive dataset is containing mostly aesthetically pleasing photos, whereas the negative contains mostly not aesthetically pleasing photos .

\subsection{Model}
The model we propose is based on deep convolutional neural network called AlexNet~\cite{imagenet2012}. The network is made up of 5 convolutional layers, max-pooling layers, dropout layers, and 3 fully connected layers. We use ReLU activations, like in the original AlexNet. The main modification is the size of the last fully connected layer -- as we use our network to perform binary classification, we decrease the size of the last layer from 1000 to 2 neurons. We use cross entropy loss function as our optimization objective.  In our experiments, we use modified AlexNet in two scenarios. In the first one, we use this neural network as a feature extractor and combine it with a Support Vector Machine (SVM) and a random forests classifier. We treat this approach as a baseline for the second approach, where we fine-tune a pre-trained AlexNet for the purposes of aesthetic value assessment.

\section{Experiments}
\label{sec:experiments}
Dataset described in Section~\ref{sec:dataset} allows us to perform various experiments related to the researched topic. Based upon previous work~\cite{flickrpopularity2014,flickrstyle2013,eyeem2016}, we focus on convolutional neural networks as they are proven in tackling computer vision problems, like image classification. In all of our experiments we use the \textit{Caffe} framework in order to perform feature extraction or fine tuning.

In this section, we describe the experiments conducted. First part presents the baseline methods that use a convolutional neural network as a feature extractor for various classifiers. The next part describes the results obtained after fine-tuning of our convolutional neural network for the purpose of our problem. The last part contains a brief analysis of the model created to solve the problem stated in the paper.

\subsection{Baseline}
\label{sec:baseline}
As a baseline of our experiments, we use an AlexNet trained on ImageNet classification problem as a feature extractor. We perform forward propagation using images from our dataset and extract activations from the sixth layer. This is similar to the approach used in the previous work~\cite{flickrpopularity2014}. We determine the optimal hyperparameters for the SVM classifier by conducting experiments on a smaller set of data. That results in selecting the RBF kernel with $C$ parameter equal to 10 and $\gamma$ parameter equal to $10^{-6}$. The number of decision trees is set to 10 considering the RF classifier. We then train SVM and RF classifiers on the full dataset. Both of the models are implemented using the \textit{scikit-learn} library available for Python. The results are presented in Table~\ref{tab:baseline}.

\begin{table}[H]
  \begin{center}       
    \begin{tabular}{| l | l | l | l | l |}
      \hline
      Classifier & Precision & Recall & F1-score & Accuracy \\ \hline
      SVM & 0.68 & 0.68 & 0.68 & 0.675 \\ \hline
      RF & 0.61 & 0.61 & 0.60 & 0.608 \\ \hline
    \end{tabular}
  \end{center}
  \caption{Results from the baseline experiments.} 
  \label{tab:baseline}
\end{table}

\subsection{Fine-tuning AlexNet for aesthetic value assessment}
\begin{figure}[H]
  \begin{center}
    \includegraphics[width=0.5\textwidth]{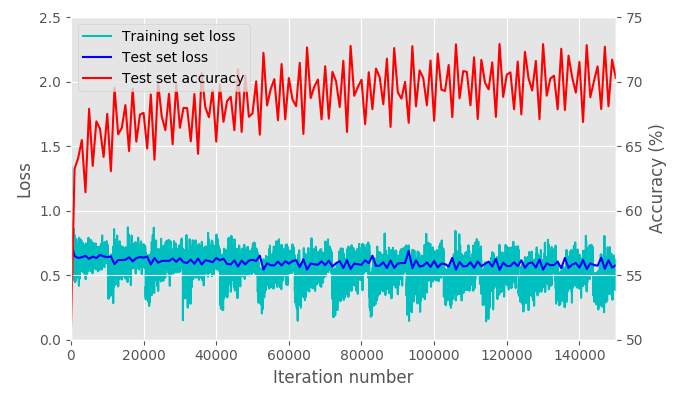}
  \end{center}
  \vspace{-3mm}
  \caption{\label{fig:alexnet} Learning curves for the training process with the best hyperparameter values. Red curve represents accuracy on the test set, whereas green and blue curves represent the training set loss and the test set loss respectively.}
\end{figure}

After defining the baseline we conduct multiple experiments using fine-tuning approach. Like in the previous experiments, we use the AlexNet convolutional neural network. We define the last layer as containing two neurons. During training, we use stochastic gradient descent solver. As we fine-tune the network, starting from the weights learned on \textit{ImageNet} dataset, we use smaller learning rate on the fine-tuned weights and bigger learning rate on weights which are randomly initialized (the last layer of the network). After trying various values for the hyperparameters, we are able to pick optimal ones, which results in the training illustrated on Figure~\ref{fig:alexnet}.

\vspace{-2mm}
\vspace{-2mm}

We train the model for 450 000 iterations, which equals to over 43 epochs with batch size of 50 examples. The trained model achieves accuracy of 70.9\%. The comparison to the baseline is presented in Table~\ref{tab:comp}.

\begin{table}[H]
  \begin{center}       
    \begin{tabular}{| l | l |}
      \hline
      Method & Accuracy on the test set \\ \hline
      Baseline (SVM) & 0.675 \\ \hline
      Baseline (RF) & 0.608 \\ \hline
      Our method & 0.709 \\ \hline
    \end{tabular}
  \end{center}
  \caption{Comparison of our method to the baseline.} 
  \label{tab:comp}
\end{table}
\vspace{-3mm}

Comparing our method to the baselines, we notice an improvement of \textbf{3.4 percentage points} over SVM classifier and \textbf{10.1 percentage points} over RF classifier.

\subsection{Analysis}
In order to fully understand at which features of the photos the model is looking while performing the inference, we sort the test set by the output values of the second neuron in the last layer. This value equals to our photo aesthetics score, i.e. probability that the input photo belongs to aesthetically pleasing class, according to our model. After that, we can select the best 100 (with the highest aesthetics score) and worst 100 photos (with the lowest aesthetics score), according to the model. That allows us to analyze which properties seem important to the model.

\begin{figure}[H]
  \begin{center}
    \begin{subfigure}[b]{0.4\textwidth}
      \includegraphics[width=\textwidth]{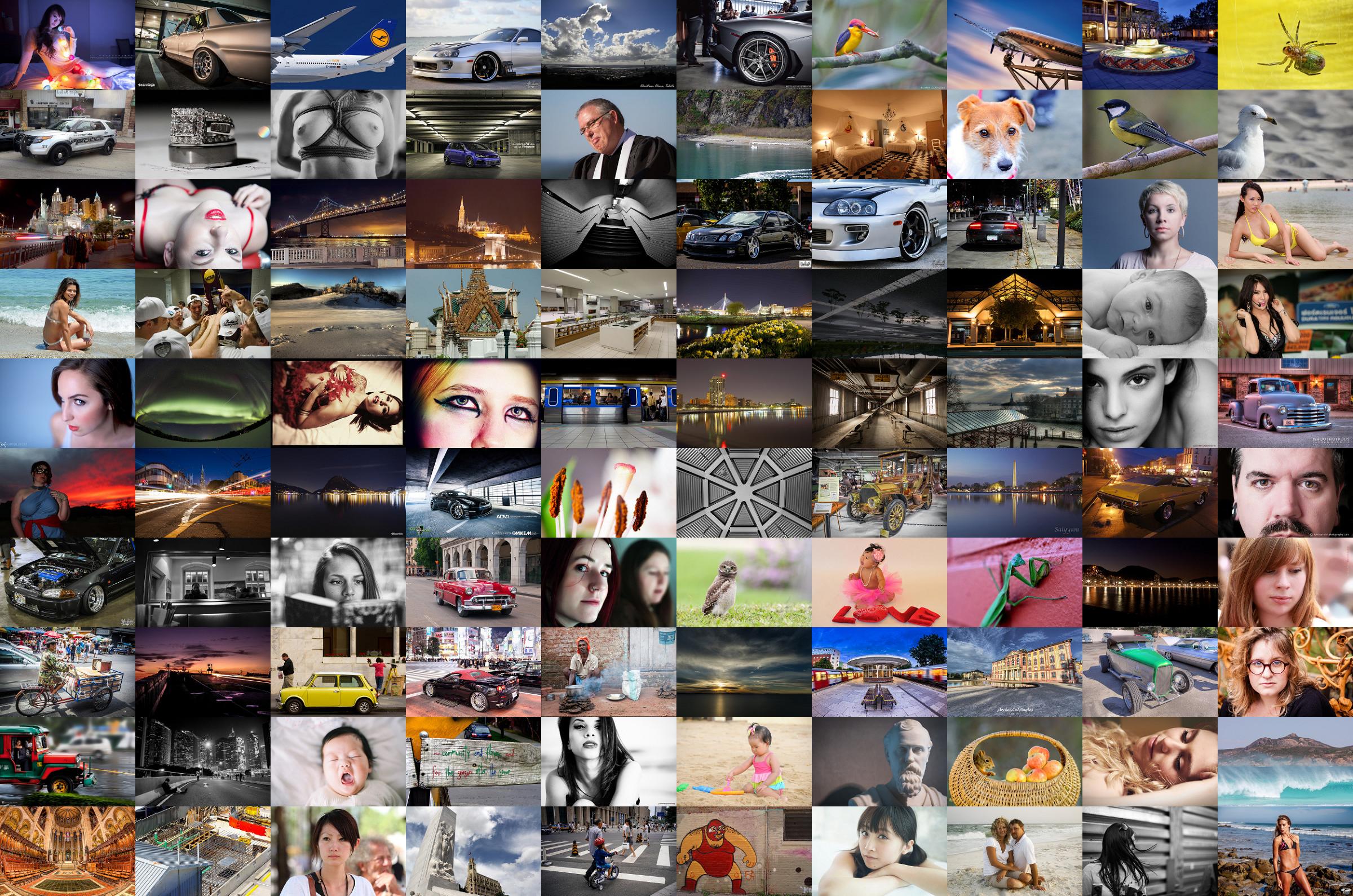}
      \caption{100 best photos according to our model.}
    \end{subfigure}
  \end{center}
  \begin{center}
    \begin{subfigure}[b]{0.4\textwidth}
      \includegraphics[width=\textwidth]{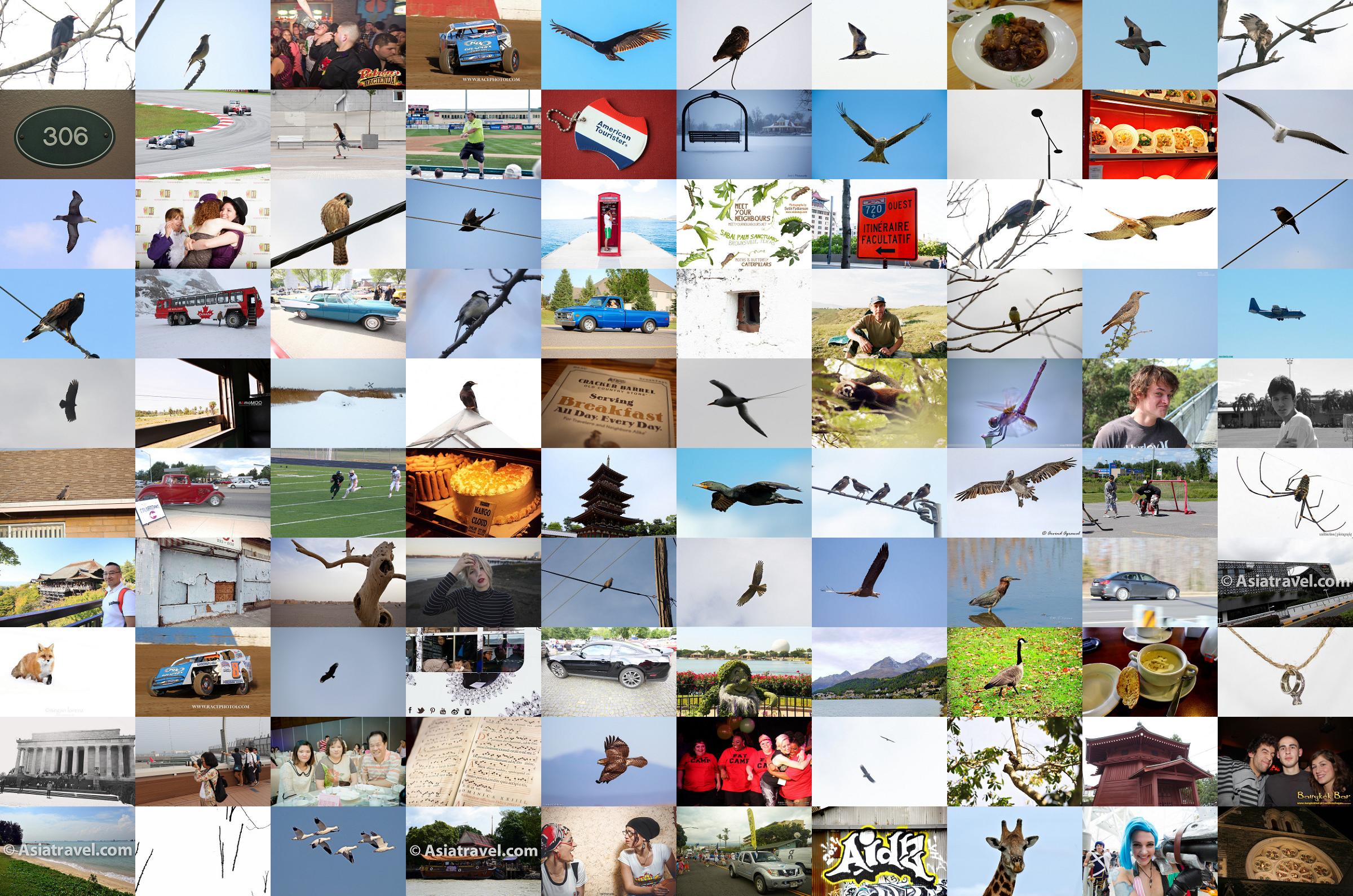}
      \caption{100 worst photos according to our model.}
    \end{subfigure}
  \end{center}
  \caption{Analysis of the network output. Two sets of 100 thumbnails with the (a)~highest and (b)~lowest score after inference using our model.}
\end{figure} 

Looking at the sorted photos, we can easily distinguish different properties of the photos that result in classifying the photo as aesthetically pleasing or not. The model is classifying the photo as a~good one, when it has the following properties:

\begin{itemize}
  \item saturated colors,
  \item high sharpness -- at least in some part of the photo,
  \item  main subject standing out from the background,
  \item high contrast.
\end{itemize}

On the other hand, these properties of the photo are likely resulting in classifying it as not aesthetically pleasing:
\begin{itemize}
  \item flat backgrounds,
  \item small main object,
  \item low contrast,
  \item wrong white balance,
  \item poor lightning or exposure.
\end{itemize}

\section{Implementation}
\label{sec:implementation}
This chapter describes the implementation of the application that was the main aim of this paper. It starts with the schema of the architecture, followed by a brief description of the building blocks of the application.

The system is deployed as an web application, in order to ensure that the application is easily available to any user without problematic installation. 

\subsection{Architecture}
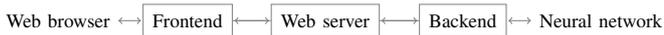
\begin{figure}[H]
  \begin{center}
    \resizebox{0.5\textwidth}{!}{%
      \begin{tikzpicture}[shorten >=1pt,->,draw=black!50, node distance=\layersep]
        \tikzstyle{every pin edge}=[<-,shorten <=1pt]
        \tikzstyle{module}=[rectangle,draw,minimum size=17pt,inner sep=5pt]

        \node[module, pin={[pin edge={<->}]left:Web browser}] (frontend) at (0,-1) {Frontend};
        \node[module] (webserver) at (2.5,-1) {Web server};
        \path (frontend) edge (webserver);
        \path (webserver) edge (frontend);
        \node[module, pin={[pin edge={<->}]right:Neural network}] (backend) at (5,-1) {Backend};
        \path (webserver) edge (backend);
        \path (backend) edge (webserver);
      \end{tikzpicture}
    }%
  \end{center}
  \caption{Architecture of our system.}
  \label{fig:architecture}
\end{figure} 

\subsection{Frontend and web server}
Both of these layers are hosted on Heroku platform. Heroku is a cloud Platform-as-a-Service (PaaS) supporting several programming languages. Concretely, we use Python in our case to create the web server. It is implemented using Flask microframework. The web server serves a minimalistic webpage to the user, along with the frontend JavaScript code that runs in the browser.

The interface allows the user to upload any photo for analysis using our system. The image is uploaded using HTML5 File API. After the upload, the photo is displayed on the webpage and resized to smaller size if necessary. The resize process is done in JavaScript, fully in the browser. That saves some bandwith and speeds up the process, as the photo needs to be sent to the machine learning backend, which is remote. The photo is sent with AJAX call to the Flask web server, which in turn uses ZeroMQ library -- exactly its socket implementation -- to send the photo to our machine learning backend for processing. The backend responds with a aesthetics score for the photo that was sent.

\subsection{Backend and neural network model}
The backend is also implemented using Python and deployed on a machine located at the Faculty of Electronics and Information Technology at Warsaw University of Technology. The machine is equipped with a NVIDIA GeForce 780 Ti GPU which is used during the inference process. The same GPU was used to train the neural network that we use. For communication with the Flask web server we use ZeroMQ library, which is sending data through an SSH tunnel.

The backend application is using Python bindings for the Caffe library to create AlexNet neural network in GPU memory. The weights for the deployed model are loaded during startup from a file. After receiving an image, the backend performs inference on a GPU using our model and returns a photo aesthetics score via ZeroMQ.

\section{Conclusions}
\label{sec:conclusions}
In this paper, we analyzed what makes photos aesthetically pleasing to the viewers. Specifically, we proposed a method to create a classifier that scores a photograph basing on its aesthetic value using only visual cues. The classifier itself could have numerous applications, which are listed in the introduction of this paper. However, there is also another benefit that comes from the analysis of the results. Namely, the convolutional neural network trained on our dataset provided us with some interesting remarks, which helped us to understand what makes a photograph aesthetically pleasing to the viewer.

However, our paper provides also future research opportunities. One could for example try to investigate if the neural network will perform better if we provide higher resolution photographs to its inputs. Other than that, future researchers may try to use our dataset to train more sophisticated deep convolutional neural networks. Recent architectures, like for example a \textit{Residual Network (ResNet)} developed by \textit{Microsoft Research}~\cite{resnet2016} or \textit{GoogleNet} created by \textit{Google}~\cite{googlenet2015} could achieve better performance. Moreover, it is easy to spot that different photography genres have different rules, e.g. landscape photography often benefits from large depth of field, whereas portrait photography is often associated with shallow depth of field. One could use this fact to train different models for different photography genres, which could bring interesting results.

\section*{Acknowledgments}
The authors would like to thank the Faculty of Electronics and Information Technology at Warsaw University of Technology for providing the necessary hardware. We gratefully acknowledge the support of NVIDIA Corporation with the donation of the Titan X Pascal GPU used for this research.

\section*{Appendix}
The web interface of the application is accessible here: \url{http://photo-critic.herokuapp.com}, whereas the dataset is available under the following links:
\begin{compactitem}
  \item training set: \url{https://goo.gl/yYw18a},
  \item test set: \url{https://goo.gl/Q7AVLZ}.
\end{compactitem}
\bibliographystyle{IEEEtran}
\bibliography{sources}

\end{document}